\documentclass{article}

\usepackage[preprint]{neurips_2024}

\usepackage[utf8]{inputenc} % allow utf-8 input
\usepackage[T1]{fontenc}    % use 8-bit T1 fonts
\usepackage{hyperref}       % hyperlinks
\usepackage{url}            % simple URL typesetting
\usepackage{booktabs}       % professional-quality tables
\usepackage{amsfonts}       % blackboard math symbols
\usepackage{nicefrac}       % compact symbols for 1/2, etc.
\usepackage{microtype}      % microtypography
\usepackage{lipsum}		% Can be removed after putting your text content
\usepackage{graphicx}
\usepackage{natbib}
\usepackage{doi}
\usepackage{subcaption}
\usepackage{amsmath}
\usepackage{enumitem}

\usepackage[textwidth=2.5cm]{todonotes}

\newcommand{\method}{HyperCloning}

\newcommand{\stackedMatrix}[4]{
    \left[\begin{array}{cc}
        #1 & #2 \\
        #3 & #4
    \end{array}\right]
}

\newcommand{\stackedVector}[2]{
    \left[\begin{array}{c}
        #1 \\
        #2
    \end{array}\right]
}

\newcommand{\transposeStackedVector}[2]{
    \left[\begin{array}{cc}
        #1 & #2
    \end{array}\right]
}

\title{Scaling Smart: Accelerating Large Language Model Pre-training with Small Model Initialization}

\author{
Mohammad Samragh\thanks{Correspondance to msamragh@apple.com and  farajtabar@apple.com.} \And Iman Mirzadeh \And Keivan Alizadeh Vahid \And Fartash Faghri \And Minsik Cho \And Moin Nabi \And Devang Naik \And Mehrdad Farajtabar$^*$  \AND Apple 
}

\begin{document}

\maketitle

\begin{abstract}
%The pre-training phase of language models often begins with randomly initialized parameters. Training this large set of random parameters can be extremely slow and costly. While small language models are less expensive to train, they often cannot achieve the accuracy of large models. In this paper, we explore an intriguing idea: what if we first train a small language model and then use it to initialize a large language model for pre-training? In other words, can we develop a method to initialize large language models using smaller pre-trained models? Will such initialization bring any benefits in terms of training time and final accuracy? 

The pre-training phase of language models often begins with randomly initialized parameters. With the current trends in scaling models, training their large number of parameters can be extremely slow and costly. In contrast, small language models are less expensive to train, but they often cannot achieve the accuracy of large models. In this paper, we explore an intriguing idea to connect these two different regimes: Can we develop a method to initialize large language models using smaller pre-trained models? Will such initialization bring any benefits in terms of training time and final accuracy? 
In this paper, we introduce \method{}, a method that can expand the parameters of a pre-trained language model to those of a larger model with increased hidden dimensions. Our method ensures that the larger model retains the functionality of the smaller model. As a result, the larger model already inherits the predictive power and accuracy of the smaller model before the training starts. We demonstrate that training such an initialized model results in significant savings in terms of GPU hours required for pre-training large language models.  
\end{abstract}

\section{Introduction}
Modern language models are very large, and training them is expensive~\citep{kaplan2020scaling, rae2021scaling,hoffmann2022training}.
%Experimenting with such models can take a long time, and it is generally risky due to the high monetary cost. 
Experimenting with such models can be time-consuming and financially burdensome due to the high monetary cost.
%Consider the training of a 12-billion-parameter model which requires approximately 72300 GPU hours for training. Assuming a cost of \$2 per GPU per hour, the total training cost would be roughly \$140,000.
For instance, training a 12-billion-parameter model requires approximately 72,000 GPU hours~\citep{biderman2023pythia}.
The total training cost from scratch can be expensive given current pricing of public cloud compute~\citep{sevilla2022compute,cottier2024rising}.
Moreover, training can fail for reasons such as improper learning rate tuning, hardware failures, or loss divergence~\citep{narayanan2021efficient,dubey2024llama}. 
Even with careful planning, robust engineering, and thorough testing to mitigate these failure risks, the monetary cost remains staggering.

While small language models are less costly to train and impose lower financial and environmental burdens during research and development, they often lack the desired level of accuracy. This situation leaves industries and businesses that prioritize performance with no choice but to scale up and utilize larger models. However, to address the prohibitive costs of training large language models from scratch, one effective strategy is to begin with a small language model and gradually expand its parameter capacity. This approach, known as model growth in contemporary literature, explores scaling up models from modest beginnings~\citep{chen2015net2net,du2024stacking}.

% Increasing the depth of small networks to build larger models has caught significant attention in recent literature. In a transformer-based network with residual blocks, each block to contributes incrementally to the hidden features without significantly altering their core content \citep{samragh2023weight}. Consequently, deeper language models can be initialized using shallower networks with fewer blocks. For instance, in \citep{du2024stacking}, researchers propose dynamically increasing the number of layers during training, exploring various strategies for layer repetition and the optimal scheduling of model growth.
% While expanding depth has received considerable focus, expanding the hidden dimension size (width) of transformers has been less explored, likely due to its inherent complexity compared to depth scaling. 

\begin{figure}
    \centering
\includegraphics[width=0.9\linewidth]{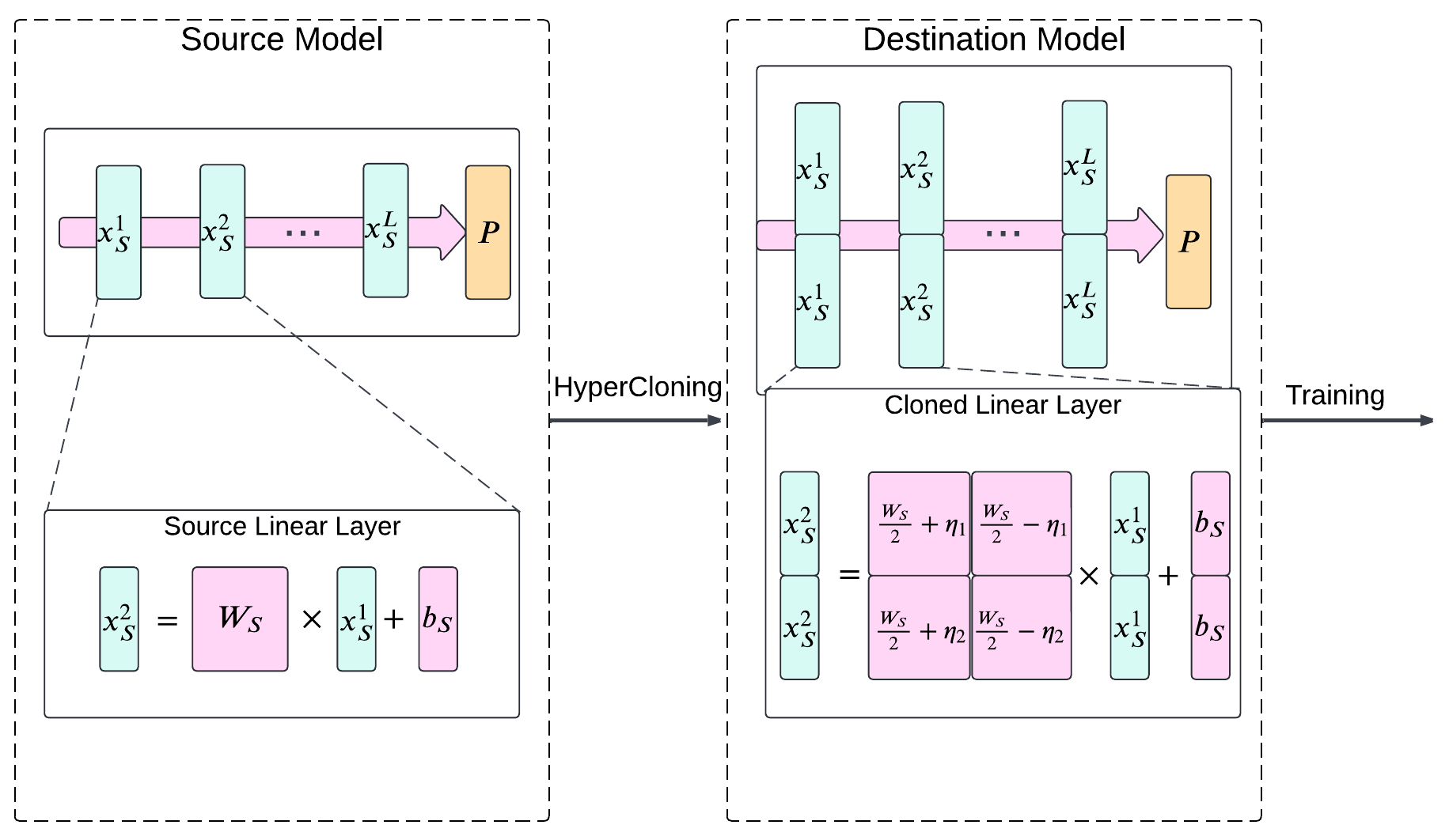}
    \caption{Illustration of \method{}. The parameters of the pretrained source network (left) are transferred to the destination network (right). In the destination model, both internal hidden representations and the final logits replicate those of the source network. This replication is achieved by precisely initializing the weights of the destination network's linear layers with the weights from the source network's linear layers, as depicted in the figure. Following this initialization, the destination network undergoes standard language model training. This initialization method enhances both the training speed and the final accuracy of the destination network.}
    \label{fig:hypercloning}
\end{figure}

In this paper, we develop a method called \method{} to increase the hidden dimensions of transformer models, enabling the initialization of larger language models from smaller ones as depicted in  Figure~\ref{fig:hypercloning}. Our method ensures a function-preserving transformation, where the output logits of the initialized model precisely match those of the smaller model. This functional preservation is advantageous as the larger language model achieves the same accuracy as the smaller model at the beginning of training. 
% compared to depth modification techniques, which may alter the model's output characteristics. With \method{}, the larger language model achieves the same accuracy as the smaller model.
And further training enhances the accuracy of the large language model.

\begin{figure}[t]
    \centering
    \begin{subfigure}[b]{0.33\textwidth}
        \centering
         \includegraphics[width=\textwidth]{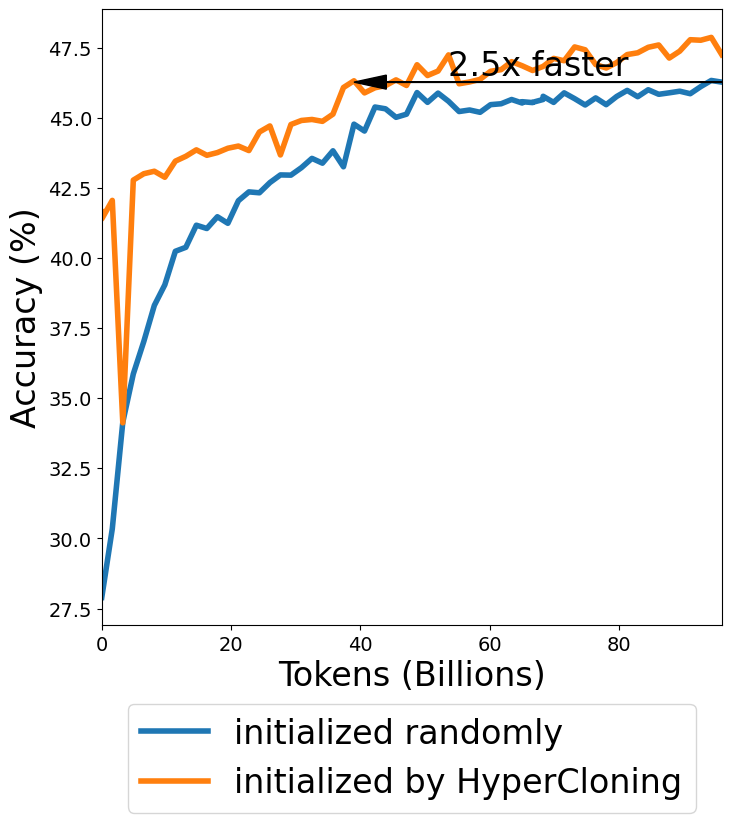}
        \caption{OPT (1.3B)}
    \end{subfigure}
    \hfill
    \begin{subfigure}[b]{0.32\textwidth}
        \centering
        \includegraphics[width=\textwidth]{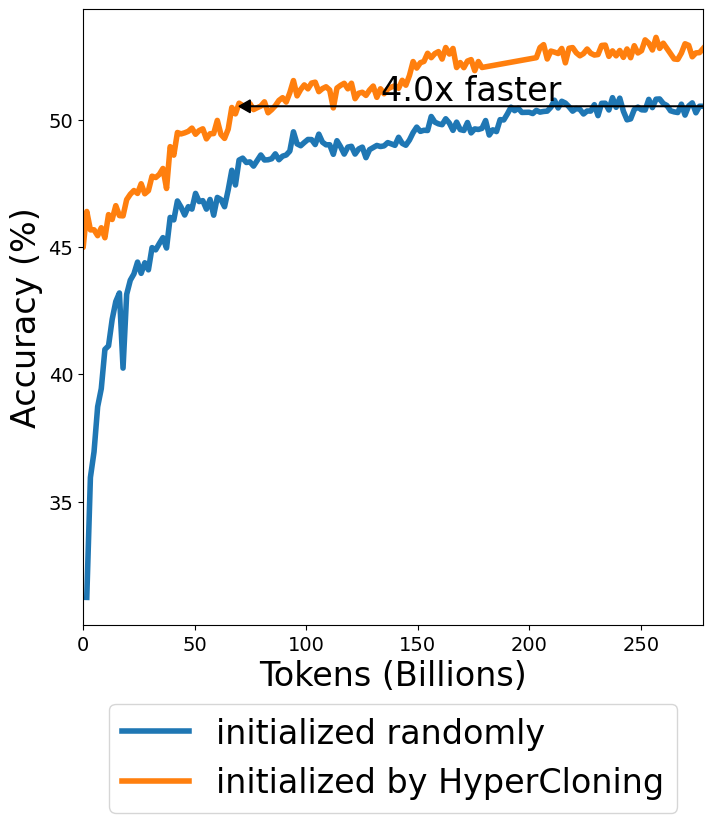}
        \caption{Pythia (1.4B)}
    \end{subfigure}
    \hfill
    \begin{subfigure}[b]{0.33\textwidth}
        \centering
        \includegraphics[width=\textwidth]{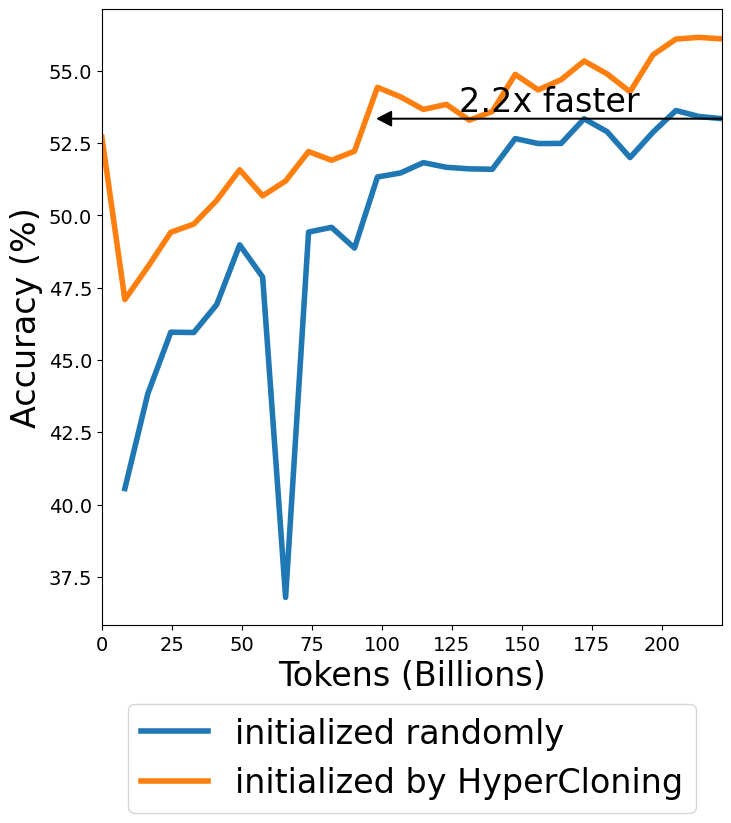}
        \caption{OLMO (2.9B)}
    \end{subfigure}
    \caption{Benchmark accuracies (averaged over 10 tasks) when models are  initialized with random weights and \method{}. Details are provided in the subsequent sections. }
    \label{fig:accs}
\end{figure}

Our experiments show that \method{} enhances both training speed and final accuracy (given a finite and reasonable training budget) compared to the classic random initialization. We evaluate our method across three families of open-source language models, namely, OPT~\citep{zhang2023opt}, Pythia~\citep{biderman2023pythia} and OLMO~\citep{groeneveld267365485olmo}, summarizing the accuracy improvements and training speed gains in Figure~\ref{fig:accs}.

% \section{Motivation and Insights}
% 

\section{Methodology}

Our goal is to design an oracle called \method{} that transfers the knowledge from a small pretrained language model to a larger model that requires training. To ensure the effectiveness of \method{}, we established several design goals:

\begin{itemize}[leftmargin=*]
\item \textbf{Expansion Dimension:} The larger network should have larger hidden dimensions compared to the smaller network, while maintaining the same number of layers in both networks.
\item \textbf{Function Preservation:} After converting the smaller model to its equivalent larger model, the logits in the final layers of both networks should match.
\item \textbf{Low Compute Overhead:} The conversion process from the smaller model to the larger model should be straightforward, avoiding heavy computations or iterative updates.
\item \textbf{Unchanged Training Loop:} For ease of deployment, the training loop should remain unchanged. The only modification should be in the network initialization.
\end{itemize}

%We targeted the first criterion to differentiate our approach from existing network growth literature by focusing on expanding network width rather than depth. 
%\mf{In this sentence cite a few of them that do depth expansions while contrasting yourself.}
In contrast to the mainstream model expansion approaches that increase the depth~\citep{gong2019efficient,samragh2023weight,yang2020progressively,karp2024landscape, li2023flm, wang2023learning}, the first criteria targets a complementary techniques that can be accompanied by any of these methods to provide a full recipe for model scaling. Width scaling can be beneficial for increased model accuracy, robustness, and inference efficiency, compared to solely increasing depth.
The second criterion gives the model a warm-start by ensuring that the larger model performs at least as well as the pretrained smaller model in the begining of training, leading to faster convergence and better final accuracy.
As we'll see, our approach, also satisfies the third and fourth criteria, which are essential for maintaining efficiency and facilitating adoption in LLM training. These differentiate \method{} with expansions methods that use techniques such as distillation to transfer knowledge~\citep{xu2024survey,zhong2023seeking}, as they usually require changing the training setup. 
% In the next section, we elaborate on our method for achieving these criteria.

\begin{figure}
    \centering
    \includegraphics[width=0.65\linewidth]{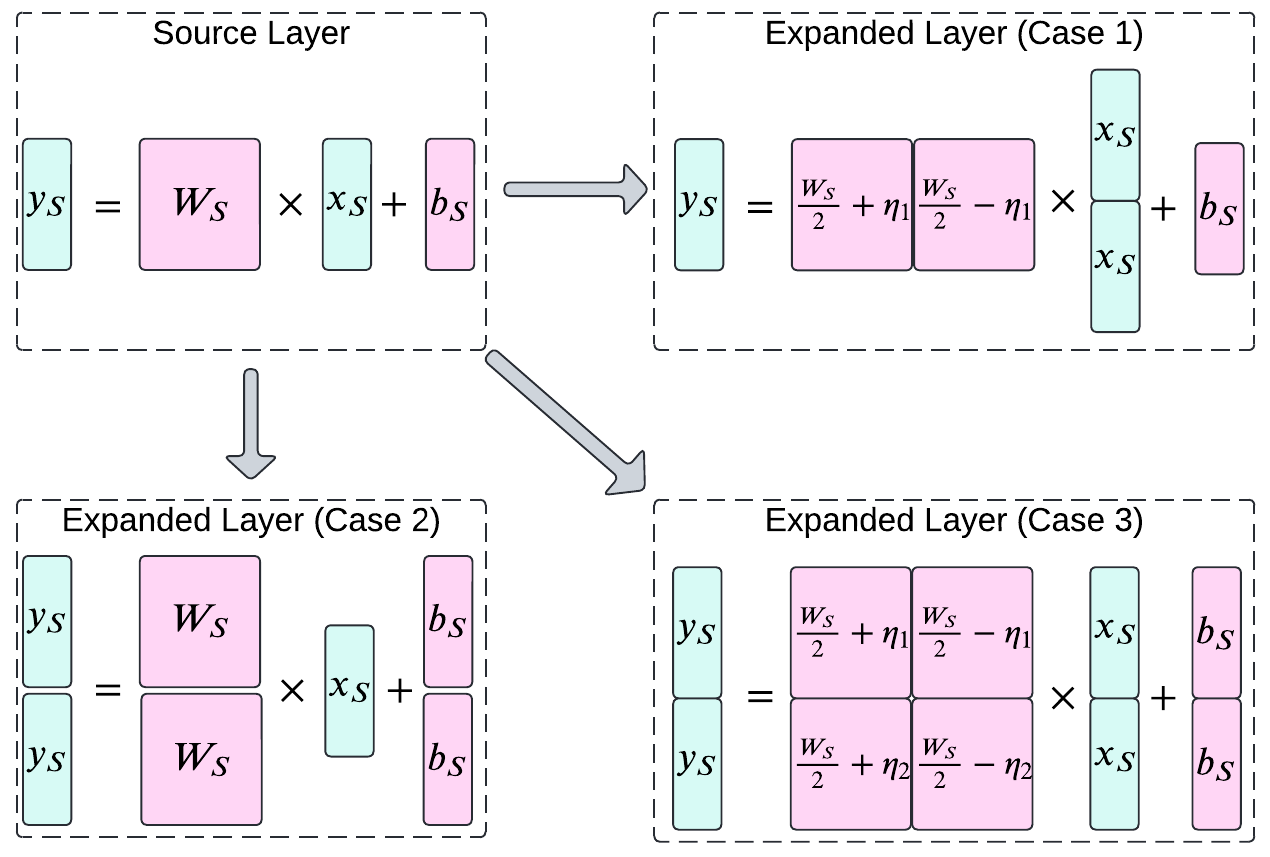}
    \caption{Demonstration of Linear layer cloning with $2$-fold expansion, where $W_s$ is the source model weight and $\eta$ is a random noise matrix.}
    \label{fig:linear_cloning}
\end{figure}

{\noindent \bf Vector Cloning.} Let $x_S \in \mathbb{R}^d$ be a hidden representation in the source (small) network. We achieve $x_D\in\mathbb{R}^{nd}$, the $n$-fold cloned version of $x_S$, by stacking $n$ copies of $x_S$ and denote it as 
$x_D =
    \left[
    \begin{array}{c}
        x_S, \ldots,x_S
    \end{array}
    \right]^\top
$.
% $$x_D =
%     \left[
%     \begin{array}{c}
%         x_S \\
%         \vdots \\
%         x_S
%     \end{array}
%     \right]
% $$
The main idea of \method{} is to establish the destination (large) network such that its hidden representations are cloned versions of the source (small) network. 
% This is depicted in Figure~\ref{fig:vector_cloning}.
% \begin{figure}[h]
%     \centering
%     \includegraphics[width=0.5\linewidth]{figures/vector_cloning.pdf}
%     \caption{Illustration of hidden vector representations in the small pretrained network (top) and the 2-fold cloned network (bottom).}
%     \label{fig:vector_cloning}
% \end{figure}
% Now that we have defined vector cloning, let us focus on common layer types and see how we can transfer parameters to enforce vector cloning. 
% For simplicity, we assume 2-fold cloning in the following description.
Consider a linear (fully connected) layer in the source network with weight parameter $W_S$ and bias parameter $b_S$. The goal of \method{} is to obtain the weight $W_D$ and bias $b_D$ in the target network such that the input and output vectors in the target model are cloned versions of those in the source network. Depending on which of the input/output dimensions are expanded, there could be three different cases shown in Figure~\ref{fig:linear_cloning}. Please refer to Appendix~\ref{app:derivations} for more specific details on the initializations for linear layers, attention layers, normalization layers, and positional embeddings. 

\section{Experiments}
\vspace{-2mm}
\noindent{\bf Model Architectures.} We perform experiments with three open-source benchmarks: OPT~\citep{zhang2023opt}, Pythia~\citep{biderman2023pythia}, and OMLO~\citep{groeneveld267365485olmo}. We choose OPT-350M, Pythia-460M, and OLMO-1B as the base pretrained models. Using \method, we then construct three larger architectures as destination networks: OPT-1.3B, Pythia-1.4B, and OLMO-2.9B. Refer to Appendix~\ref{app:training} for more information about model architectures, training dataset, and training hyperparameters.
\subsection{Results Overview}

% \begin{itemize}
%     \item talk about loss values
%     \item for OLMO show that we achieve the accuracy fast enough. Compute OLMO-7B accuracy at this many tokens. Say that we already achieve this with fewer tokens.
%     \item For Pythia and OPT compare the accuracy with the HF models. Compare the number of Tokens trained.
% \end{itemize}

\subsubsection{Comparison to Random Initialization}

In this section, we compare the training convergence of the studied models in two scenarios: (i) random initialization, which is the standard process for training language models, and (ii) initialization with \method{} from a base model. In both cases, all other hyperparameters were kept identical, including learning rate, optimizer type, number of GPU nodes, batch size, context size, and order of training data.

\begin{figure}[t]
    \centering
    \begin{subfigure}[b]{0.3\textwidth}
        \centering
         \includegraphics[width=\textwidth]{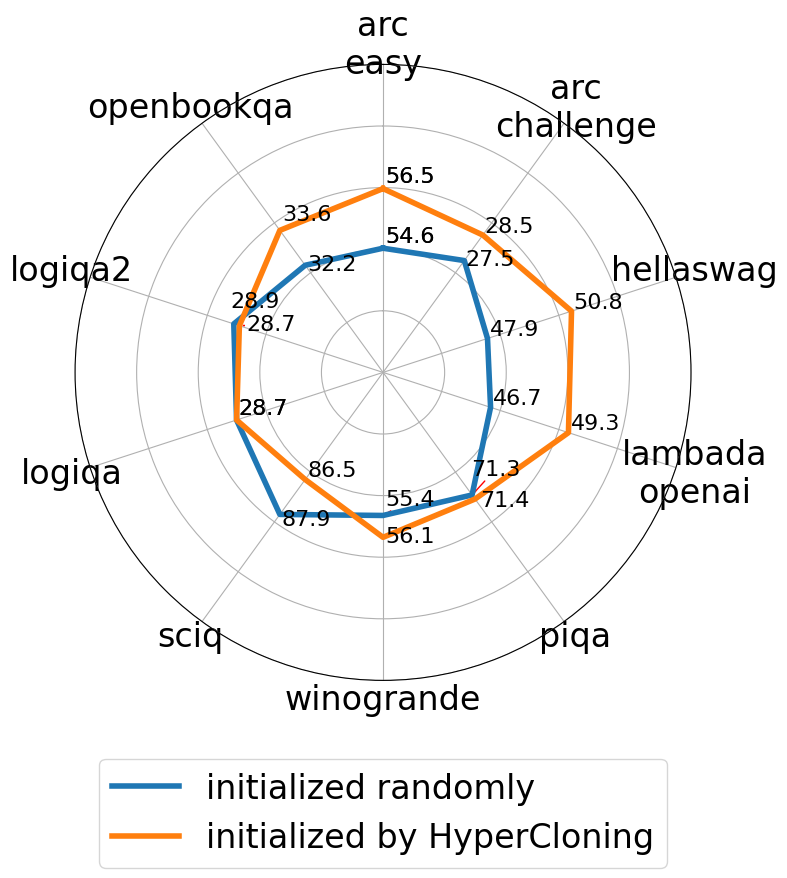}
        \caption{OPT}
    \end{subfigure}
    \hfill
    \begin{subfigure}[b]{0.3\textwidth}
        \centering
        \includegraphics[width=\textwidth]{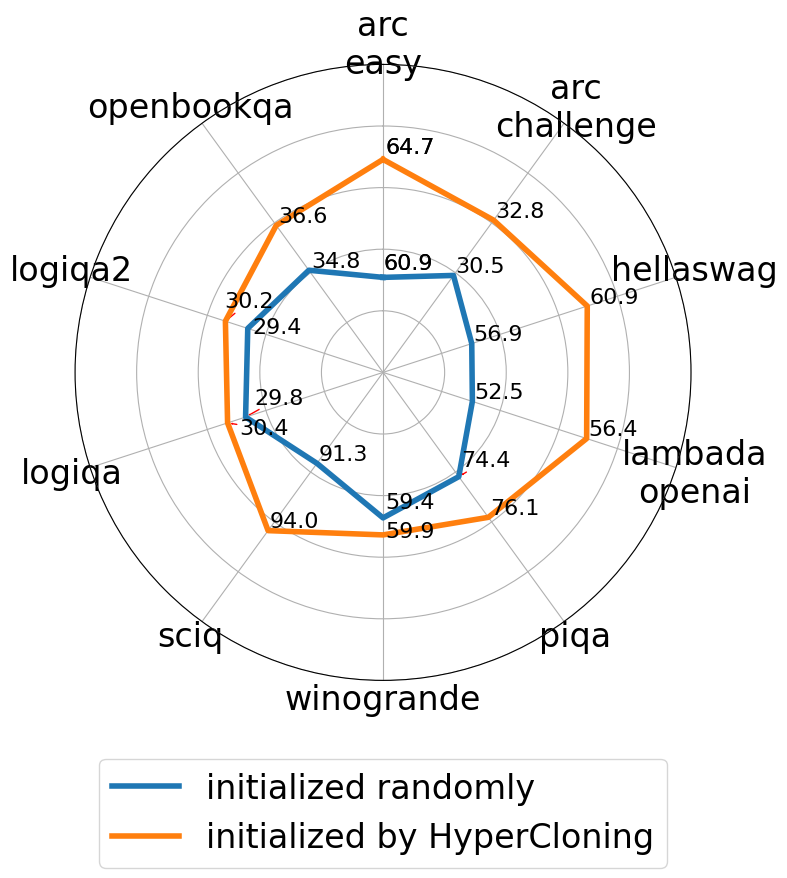}
        \caption{Pythia}
    \end{subfigure}
    \hfill
    \begin{subfigure}[b]{0.3\textwidth}
        \centering
        \includegraphics[width=\textwidth]{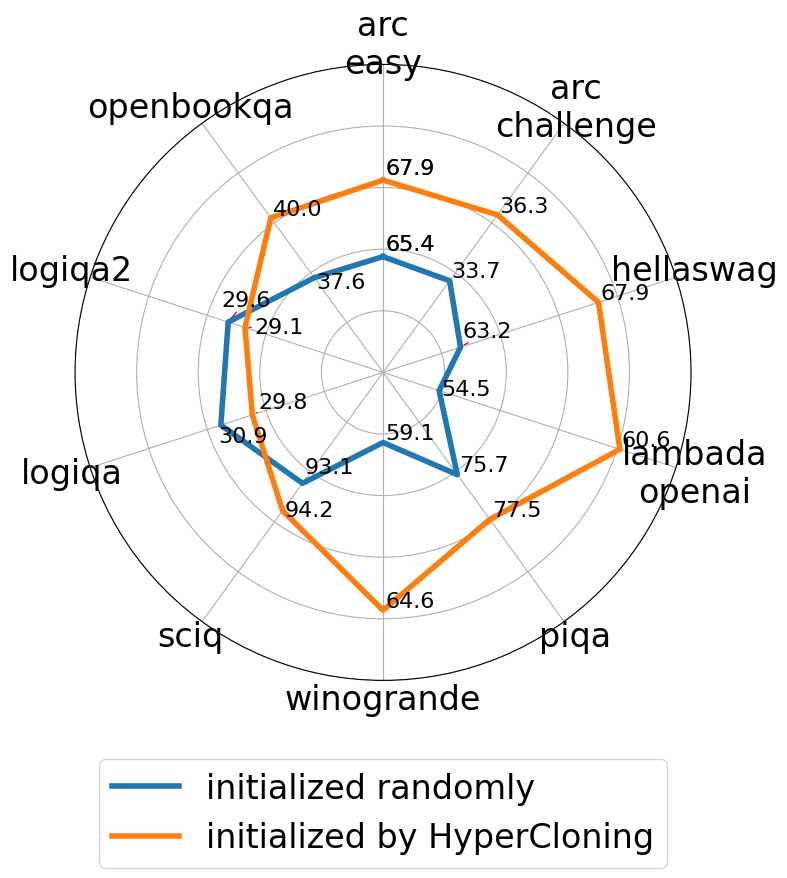}
        \caption{OLMO}
    \end{subfigure}
    \caption{Benchmark accuracies over 10 tasks when models are  initialized with random weights and \method{}.}
    \vspace{-2mm}
    \label{fig:accs-benchmarks}
\end{figure}

We compute the models' accuracy using the Harness framework~\citep{eval-harness}, an open-source and widely-used tool for LLM evaluation. Accuracies are measured over 10 different tasks, and the final accuracies for both random initialization and \method{} are presented in Figure~\ref{fig:accs-benchmarks}. As shown, \method{} significantly improves the accuracy of the models after convergence.

Additionally, we measure the average accuracy over the 10 tasks and present its trend during training in Figure~\ref{fig:accs}, which we present early in the paper. As observed, \method{} enables the network to reach the final accuracy of the random initialization baseline much faster, with a speedup ranging from 2.2x to 4x across different model types. The better final accuracy and faster convergence achieved by \method{} can be attributed to the transfer of knowledge from the base model. For example, the base model for the OLMO architecture was already trained on 2.4T tokens, and this knowledge was transferred to our model before training started. Note that the base models are freely available; we simply downloaded them from HuggingFace. In practice, \method{} can leverage previously trained models, thus offering a cost-saving advantage. Consequently, the model initialized with \method{} begins with high accuracy and can converge to a better solution with significantly fewer training tokens (i.e., 250B tokens rather than 2.4T).

One notable observation is that models initialized with \method{} tend to exhibit catastrophic forgetting at the beginning of training. This is evident in the training curve for the OLMO benchmark. However, our experiments show that with sufficient training, this forgetting can be compensated for. Despite the initial catastrophic forgetting, \method{} still outperforms random initialization by a large margin. Understanding the underlying causes of catastrophic forgetting, identifying strategies to mitigate it, and exploring why \method{} continues to outperform random initialization despite its occurrence are valuable avenues for future research. We believe these areas hold great potential for further enhancing our method.

\subsection{Analyzing \method{}: Weight Symmetry}

For an $n$-fold cloning, the target weights in the target network are initialized with blocks of source weights normalized by $n$. Consequently, the weights in the target network have a standard deviation that is $\frac{1}{n}$ of the standard deviation of the source network weights. This approach aligns with the standard deviation requirement proposed by \citep{glorot2010understanding} and offers benefits over existing methods like those in~\citep{wang2023lemon} and~\citep{chen2015net2net}.

However, our method, \method{}, initializes parts of the weight parameters as duplicates of each other. As noted by~\citep{wang2023lemon}, this duplication raises concerns that the duplicated neurons or weights might not learn independently, potentially limiting the model's capacity to utilize all parameters effectively. Nonetheless, we observe that this issue does not occur in our implementation, likely due to the randomness introduced by techniques such as dropout.

To analyze the evolution of these weight patterns during training, we define a metric to assess the symmetry in a cloned matrix. In the 2-fold cloning scenario depicted in Figure~\ref{fig:linear_cloning}, case 3, each row of the matrix contains two identical horizontal vectors. We measure the cosine similarity between these vectors for each row and calculate the average cosine similarity across all rows. This metric provides an indication of the similarity between the vectors in the matrix.

\begin{figure}[t]
    \centering
    % First row of subplots
    \begin{subfigure}[b]{0.3\textwidth}
        \centering
         \includegraphics[width=\textwidth]{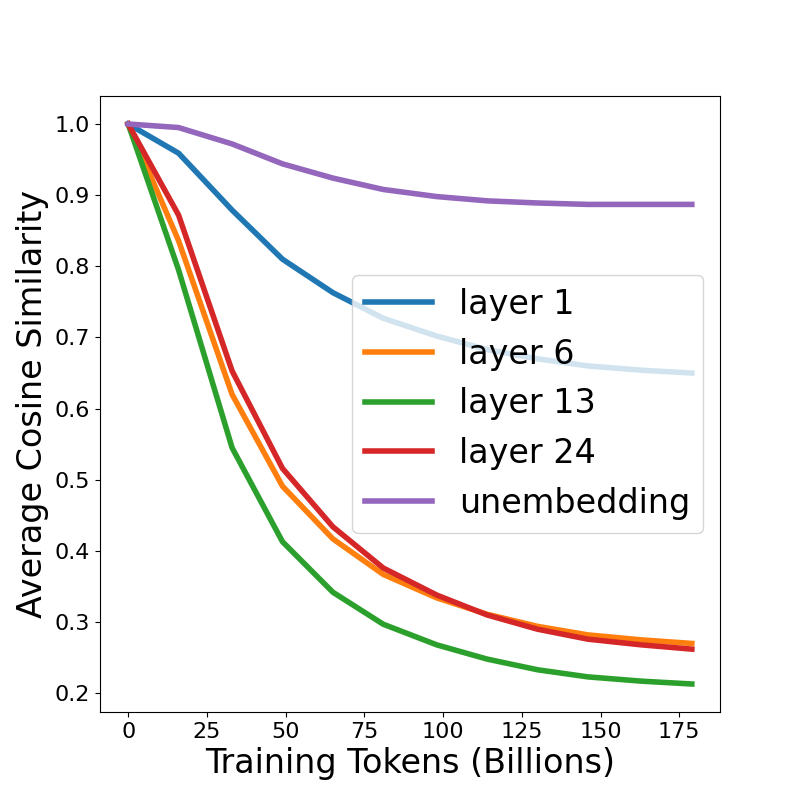}
        \caption{OPT}
    \end{subfigure}
    \hfill
    \begin{subfigure}[b]{0.3\textwidth}
        \centering
        \includegraphics[width=\textwidth]{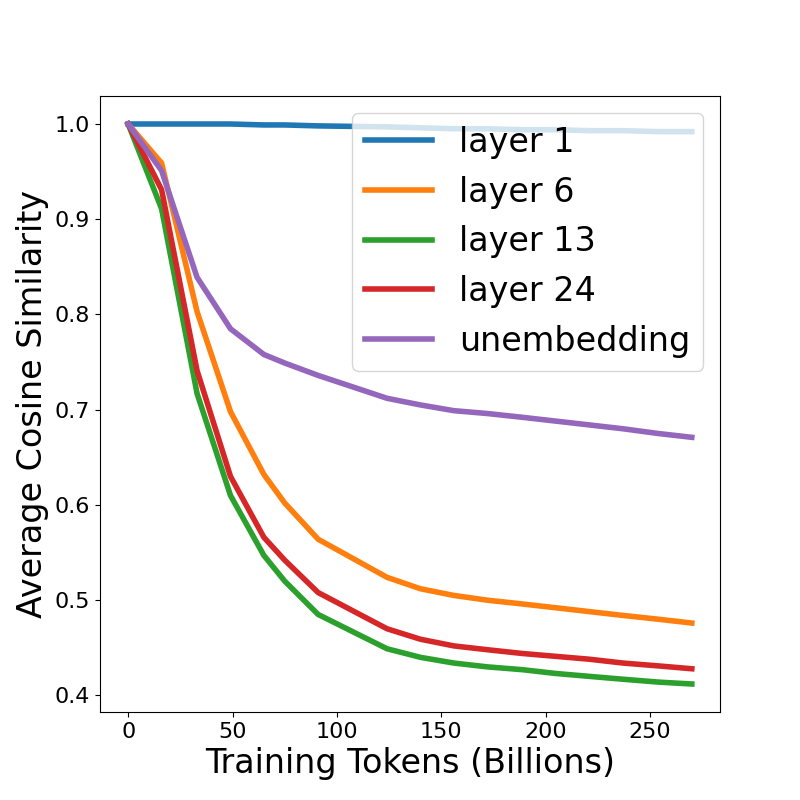}
        \caption{Pythia}
    \end{subfigure}
    \hfill
    \begin{subfigure}[b]{0.3\textwidth}
        \centering
        \includegraphics[width=\textwidth]{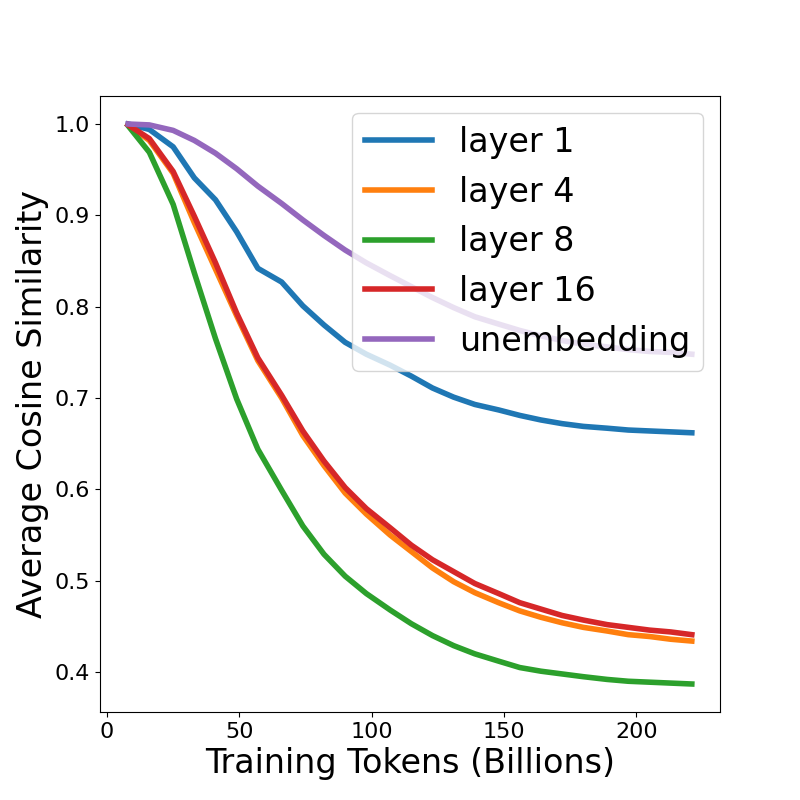}
        \caption{OLMO}
    \end{subfigure}
    \caption{Evolution of average cosine similarities during training of the target network at up-project feed forward layers and the final unembedding layer.}
    \label{fig:cosine}
\end{figure}

Figure~\ref{fig:cosine} shows the evolution of cosine similarities for several selected layers in our studied networks. Initially, the cosine similarities of all layers are 1, showing a complete symmetry in the weights. As training progresses, we observe that the cosine similarity decays in most layers. This suggests that the model is utilizing its effective parameter space during training. While this analysis provides insights into the evolution of model weights, further studies are worthwhile in the future.
%To save space and focus on the key results, we believe this level of elaboration is sufficient for this paper.

\subsection{Analyzing \method{}: Principal Components}

Another way to analyze the convergence of \method{} is by examining the ranks of the weight matrices. Consider the weight matrices shown in Figure~\ref{fig:linear_cloning}. Due to the replicating nature of our cloning algorithm, it is evident that the rank of the cloned matrix is at most equal to the rank of the base matrix. Essentially, the rank of the cloned matrix is half of its maximum possible value at initialization. This implies that, while the model has reasonable accuracy at initialization, it is not fully utilizing its capacity for making predictions. The concern is that the model might continue underutilizing this capacity even after training is completed. We demonstrate that this does not occur.

\begin{figure}[t]
    \centering
    \begin{subfigure}[b]{0.3\textwidth}
        \centering
         \includegraphics[width=\textwidth]{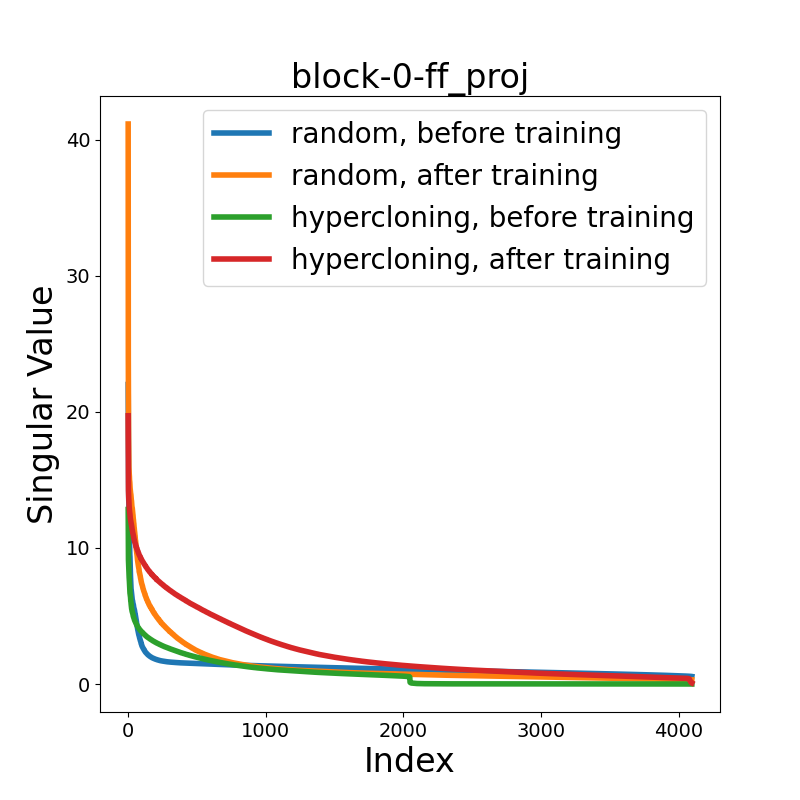}
        \caption{Block 0, up-project weights}
    \end{subfigure}
    \hfill
    \begin{subfigure}[b]{0.3\textwidth}
        \centering
        \includegraphics[width=\textwidth]{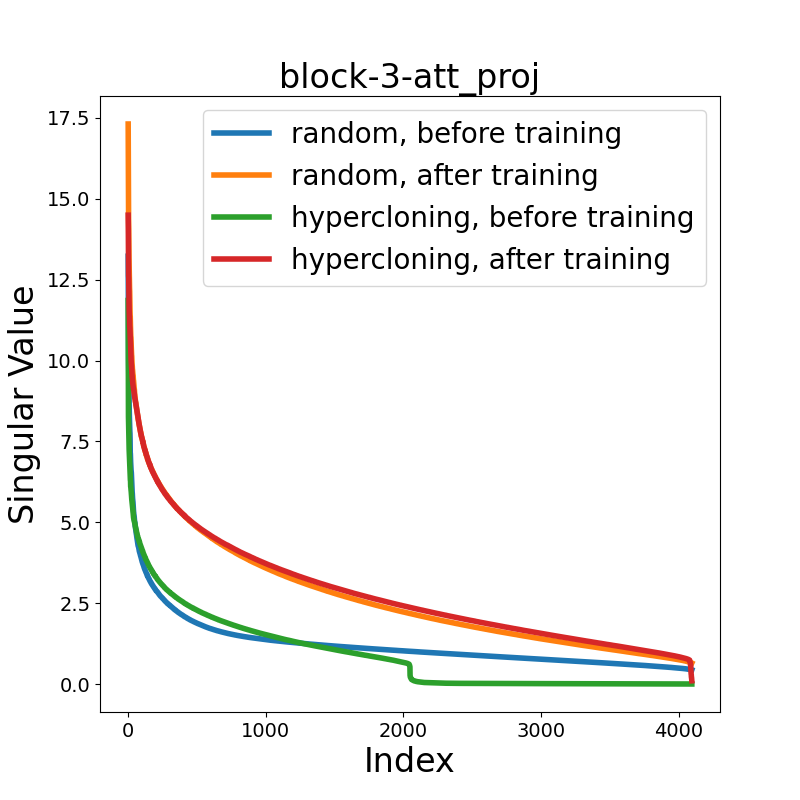}
        \caption{Block 3, QKV weights}
    \end{subfigure}
    \hfill
    \begin{subfigure}[b]{0.3\textwidth}
        \centering
        \includegraphics[width=\textwidth]{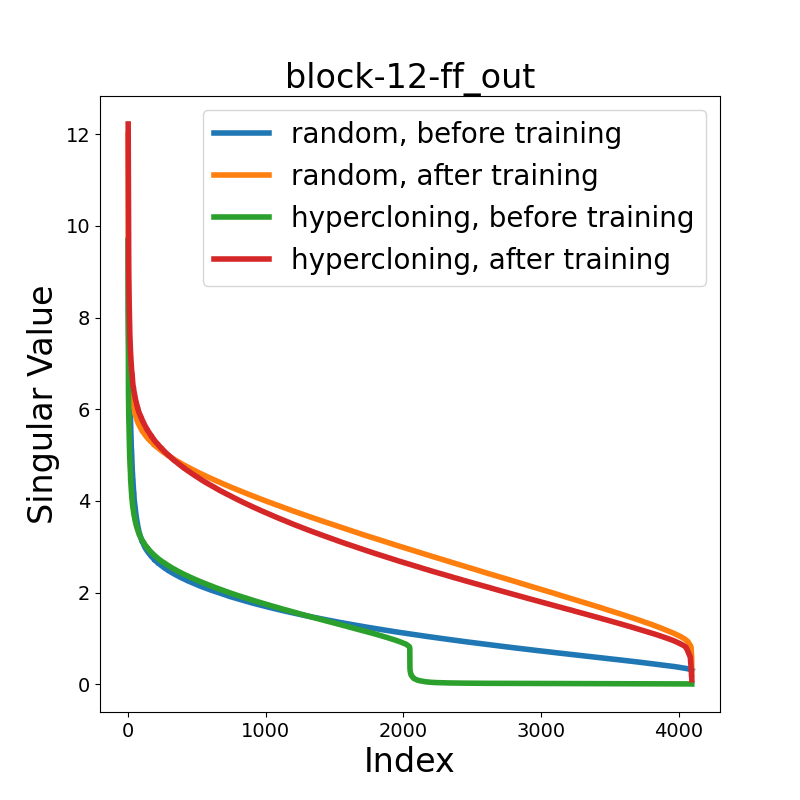}
        \caption{Block 12, down-project weights}
    \end{subfigure}
    \caption{Singular values of weights at different layers of OLMO-2.9B model.}
    \label{fig:ranks}
\end{figure}

In Figure~\ref{fig:ranks}, we show the eigenvalues of several weight matrices within our OLMO-2.9B model before and after training, for both the randomly initialized model and the model initialized with \method{}. It can be seen that half of the singular values of the \textit{before training} model initialized with \method{} are zero, whereas the randomly initialized model does not exhibit this behavior. However, after training, the model initialized with \method{} achieves similar high-rank weights to those in the randomly initialized model.

\subsection{Alternative Expansion Methods}
In our original formulation for the expanded weights, we proposed $W_L = \stackedMatrix{\frac{W_S}{2}}{\frac{W_S}{2}}{\frac{W_S}{2}}{\frac{W_S}{2}}$. However, this is not the only weight parameter configuration that can satisfy function preservation. In this part of our analysis, we empirically evaluate several strategies for initializing $W_L$ as follows:

\begin{itemize}
    \item \textbf{Symmetric}: $W_L = \stackedMatrix{\frac{W_S}{2}}{\frac{W_S}{2}}{\frac{W_S}{2}}{\frac{W_S}{2}}$.
    %We refer to this as \textit{symmetric} initialization.
    \item \textbf{Diagonal}: $W_L = \stackedMatrix{W_S}{0}{0}{W_S}$.
    %We refer to this as \textit{diagonal} initialization.
    \item \textbf{Noisy symmetric}: $W_L = \stackedMatrix{\frac{W_S}{2}+\eta_1}{\frac{W_S}{2}-\eta_1}{\frac{W_S}{2}+\eta_2}{\frac{W_S}{2}-\eta_2}$, where $\eta_1$ and $\eta_2$ are random noise tensors of the same shape as $W_S$.
    %We refer to this as \textit{noisy symmetric} initialization.
    \item \textbf{Noisy diagonal}: $W_L = \stackedMatrix{W_S+\eta_1}{-\eta_1}{\eta_2}{W_S-\eta_2}$.
    %We refer to this as \textit{noisy diagonal} initialization.
\end{itemize}

\begin{figure}[!ht]
    \centering
    % First row of subplots
    \begin{subfigure}[b]{0.33\textwidth}
        \centering
         \includegraphics[width=\textwidth]{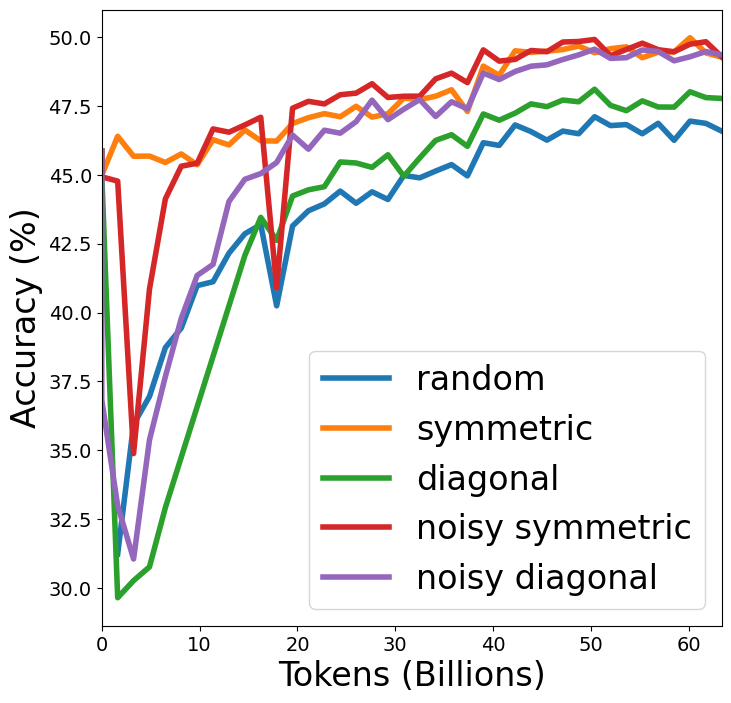}
        \caption{Average Accuracy.}
    \end{subfigure}
    \begin{subfigure}[b]{0.33\textwidth}
        \centering
        \includegraphics[width=\textwidth]{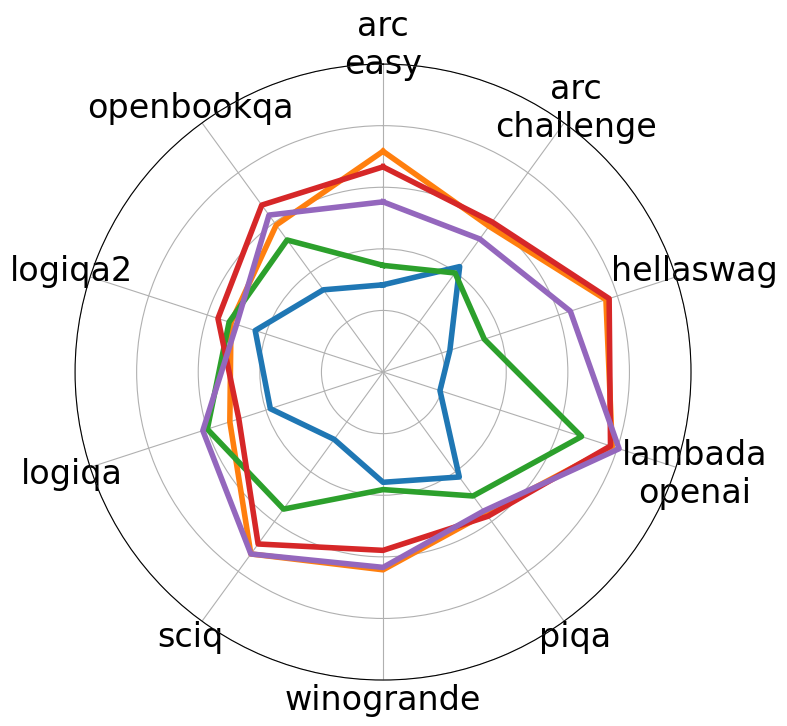}
        \caption{Benchmark Accuracies.}
    \end{subfigure}
    \caption{Effect of expanding strategy on Model Accuracy for Pythia 1.4B training.  }
    \label{fig:versions}
\end{figure}

Note that all of the above weight expansion strategies are function-preserving. Figure~\ref{fig:versions} shows the accuracy of each instantiation method. The noise values ($\eta_1$ and $\eta_2$) in these experiments are selected such that the signal-to-noise ratio is 10 dB. All cloning methods outperform random initialization. The diagonal variant achieves the smallest accuracy boost, likely due to the presence of zero values in the expanded weight matrices. The noisy diagonal version performs slightly better than diagonal; however, the symmetric and noisy symmetric methods stand out as the best. With symmetric expansion, the benefits of noise addition are minimal. Therefore, we opt for the noise-free version of the method to avoid having to tune an extra hyper-parameter, the signal-to-noise ratio.

\subsection{Effect of base model accuracy}
\begin{figure}
    \centering
    % First row of subplots
    \begin{subfigure}[b]{0.33\textwidth}
        \centering
         \includegraphics[width=\textwidth]{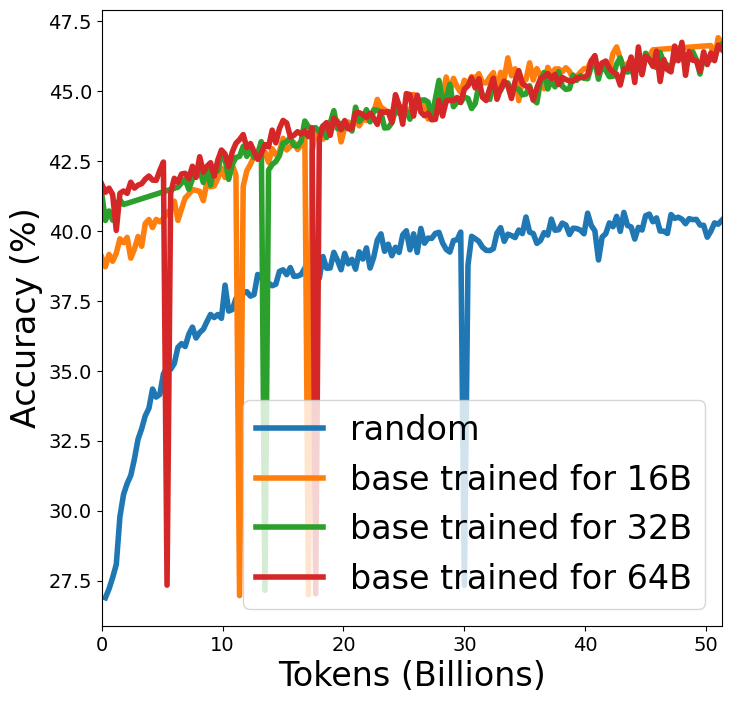}
        \caption{Average Accuracy.}
    \end{subfigure}
    \begin{subfigure}[b]{0.33\textwidth}
        \centering
        \includegraphics[width=\textwidth]{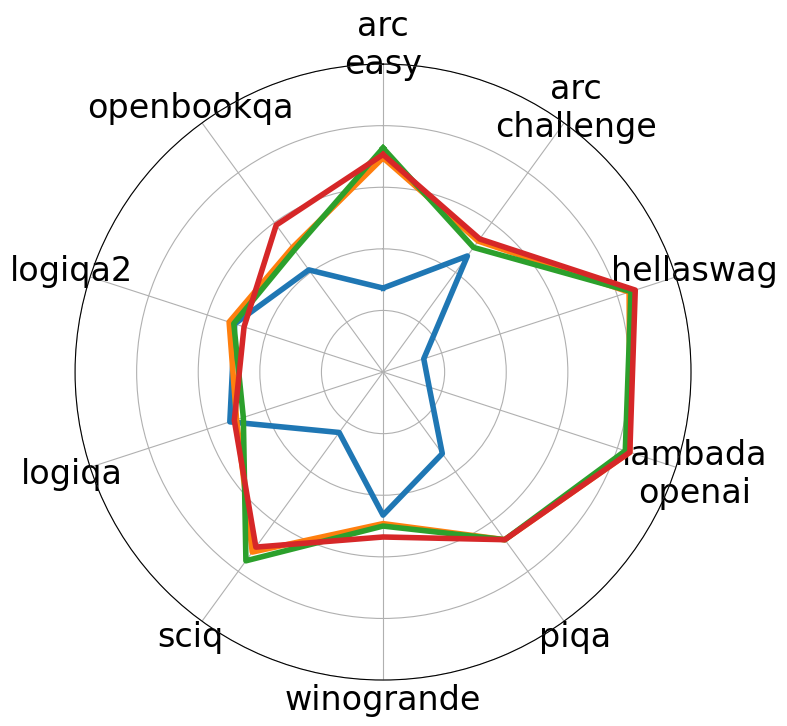}
        \caption{Benchmark Accuracies.}
    \end{subfigure}
    \caption{Effect of base model's accuracy on the convergence of target model. In this experiment, the base model is OPT-350M and the target model is OPT-1.3B.}\label{fig:base_acc_effect}
    \vspace{-0.5cm}
\end{figure}

Next, we study the effect of the base model's performance on the target model's performance. For this study, we use different checkpoints from the OPT-350M base model, trained with 16, 32, and 64 billion tokens, respectively. We initialize the target OPT-1.3B model with each of these checkpoints. Another baseline is random initialization, bringing the total number of comparison baselines to four. We observe the training convergence in Figure~\ref{fig:base_acc_effect}. As seen, initializing with the base model improves accuracy compared to random initialization when any of the base checkpoints are used for cloning. Among the cloned networks, those initialized with a more accurate base network achieve better accuracy, especially at the beginning of the training. However, as training continues, the differences between the curves become smaller.

\subsection{Effect of base model size}

\begin{figure}
    \centering
    % First row of subplots
    \begin{subfigure}[b]{0.33\textwidth}
        \centering
         \includegraphics[width=\textwidth]{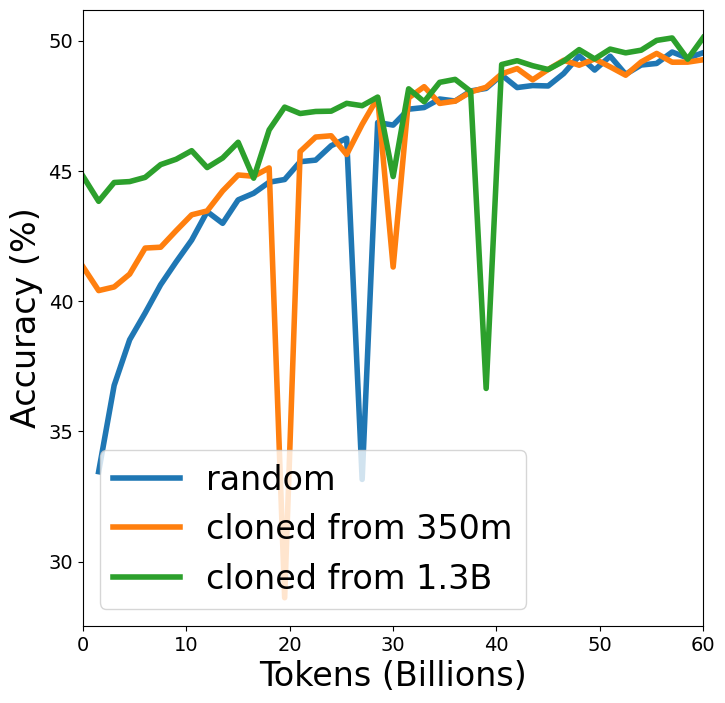}
        \caption{Average Accuracy.}
    \end{subfigure}
    \begin{subfigure}[b]{0.33\textwidth}
        \centering
        \includegraphics[width=\textwidth]{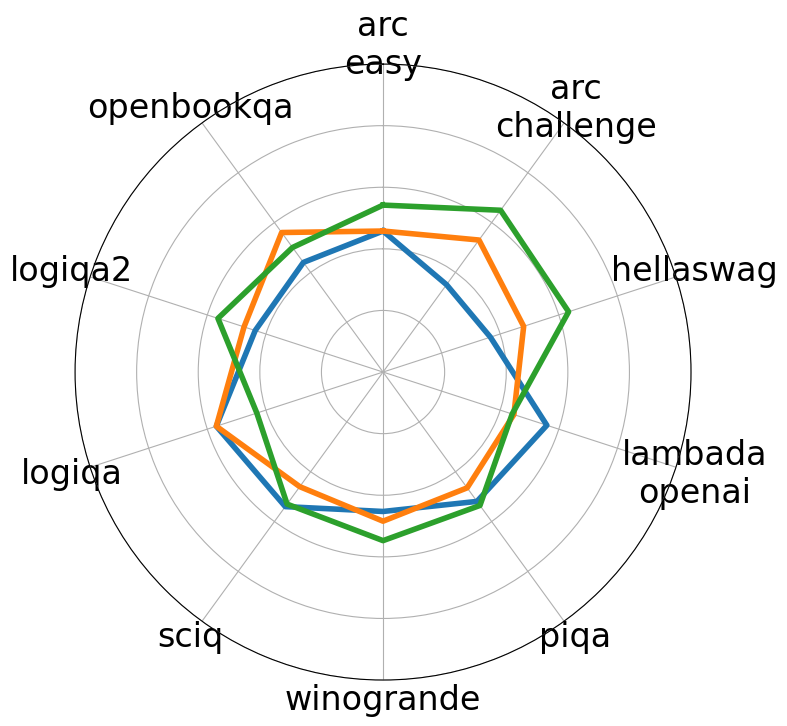}
        \caption{Benchmark Accuracies.}
    \end{subfigure}
     \caption{Effect of base model's size on the convergence of target model. In this experiment, the base model is either OPT-350M or OPT-1.3B, and the target model is OPT-5.3B.} \label{fig:base_size_effect}
\end{figure}

Next, we demonstrate the effect of the base model's size on the target network's convergence. We create the target model by doubling the hidden dimension size of OPT-1.3B, resulting in a model we call OPT-5.3B. This architecture can be initialized in two ways using \method{}: (i) with OPT-1.3B using 2-fold cloning, or (ii) with OPT-350M using 4-fold cloning. The convergence of these candidates, along with the network initialized randomly, is shown in Figure~\ref{fig:base_size_effect}. As observed, initializing with either OPT-350M or OPT-1.3B achieves faster convergence compared to random initialization, with OPT-1.3B providing better convergence than OPT-350M. This is because OPT-1.3B is larger and more accurate than OPT-350M, thereby offering a superior initialization.

\section{Related Work}

A comprehensive study on related work in the network growth literature is available in \citep{du2024stacking}, which examines various growth strategies, including depth and width expansion. Their innovative approach to depth growth involves initializing a larger model by repeating block weights. This approach is also supported by the findings of other research work~\citep{gong2019efficient,samragh2023weight,yang2020progressively,karp2024landscape, li2023flm, wang2023learning}. For instance, \citep{samragh2023weight} demonstrates that, due to the presence of residuals in transformer architectures, blocks can be removed or duplicated to achieve superior initialization compared to random methods. In terms of width growth, \citep{du2024stacking} explored several strategies, including directly copying weights, projecting weights to a larger dimension, initializing new weights to zero, and randomly initializing weights. Notably, both depth and width scaling strategies in their study do not preserve function properties. They concluded that depth growth achieves the best accuracies, while non-function-preserving width growth results in poorer performance. While their work provides valuable insights, our research takes a different direction by focusing on a function-preserving transformation in the width dimension. Further studies are necessary to fully understand the benefits of depth versus width scaling and function-preserving versus non-function-preserving transformations.

Width expansion was initially introduced by \citep{chen2015net2net} for convolutional neural networks and later explored for BERT-style transformer models in \citep{chen2021bert2bert}. Our work builds on these foundations by generalizing width expansion techniques to decoder-style transformers, which are increasingly utilized in modern large language models. Specifically, we extend the width expansion method to include attention layers, define essential cloning functions for position embeddings, and validate our approach through experiments on larger-scale models and datasets. These contributions advance the applicability of width expansion in contemporary transformer architectures.

In \citep{shen2022staged}, the authors introduce a width expansion technique where the non-diagonal elements of the expanded weight matrices are initialized to zero. Our ablation studies indicate that this diagonal initialization can lead to slower convergence compared to our symmetric initialization method. In contrast, \citep{wang2023lemon} discuss that the symmetry of neurons in an expanded network suggests these neurons may not contribute independently to the model's learning. However, our experiments demonstrate that the symmetry in weights naturally breaks during training, potentially due to random operations such as dropout.

We further propose a function-preserving noise addition mechanism to intentionally break the symmetry in weights. Our findings show that this noise addition improves the model's convergence rate. Additionally, we analyze the eigenvalues of the expanded network's weights after training and find that their distribution closely resembles that of a network trained from scratch. This result suggests that the expanded network effectively utilizes its parameter space during learning, comparable to a network trained from scratch.
\section{Conclusion}

This paper introduces \method{}, a novel initialization strategy designed to transfer weights from a smaller, pretrained source model to a larger target model. The transfer process in \method{} is straightforward, effective, and preserves the model's functionality. By using this method, we achieve faster convergence and better final accuracy during language model training. In our experiments, \method{} accelerates training by 2-4 times. Additionally, we conducted ablation studies to explore the impact of the source model's architecture and different weight-cloning techniques on the target model's convergence.

\bibliographystyle{unsrtnat}
\bibliography{references}

% \section*{References}

% References follow the acknowledgments in the camera-ready paper. Use unnumbered first-level heading for
% the references. Any choice of citation style is acceptable as long as you are
% consistent. It is permissible to reduce the font size to \verb+small+ (9 point)
% when listing the references.
% Note that the Reference section does not count towards the page limit.
% \medskip

% {
% \small

% [1] Alexander, J.A.\ \& Mozer, M.C.\ (1995) Template-based algorithms for
% connectionist rule extraction. In G.\ Tesauro, D.S.\ Touretzky and T.K.\ Leen
% (eds.), {\it Advances in Neural Information Processing Systems 7},
% pp.\ 609--616. Cambridge, MA: MIT Press.

% [2] Bower, J.M.\ \& Beeman, D.\ (1995) {\it The Book of GENESIS: Exploring
%   Realistic Neural Models with the GEneral NEural SImulation System.}  New York:
% TELOS/Springer--Verlag.

% [3] Hasselmo, M.E., Schnell, E.\ \& Barkai, E.\ (1995) Dynamics of learning and
% recall at excitatory recurrent synapses and cholinergic modulation in rat
% hippocampal region CA3. {\it Journal of Neuroscience} {\bf 15}(7):5249-5262.
% }

%%%%%%%%%%%%%%%%%%%%%%%%%%%%%%%%%%%%%%%%%%%%%%%%%%%%%%%%%%%%

\appendix

\section{Cloning Details}~\label{app:derivations}

In this section we explain the cloning process for different layer types in detail. For simplicity, we consider a 2-fold expansion of the network but the method can be generalized to a generalized $n$-fold expansion.

{\noindent \bf Cloning Linear Layers.}  In general, there can be three different expansion cases for a Linear Layer show in Figure~\ref{fig:linear_cloning}:

\begin{itemize}
    \item \textbf{Case 1:} Only the input is expanded: $x_D = \stackedVector{x_S}{x_S}$ and $y_D = y_S$. This may occur at any linear layer whose outputs are not expanded such as the unembedding layer. 
    \item \textbf{Case 2:} Only the output is expanded: $x_D = x_S$ and $y_D = \left[\begin{array}{c} y_S \\ y_S\end{array}\right]$. This may occur at any linear layer whose inputs are not expanded such as the embedding layer.
    \item \textbf{Case 3:} Both input and output are expanded : $x_D = \left[\begin{array}{c} x_S \\ x_S\end{array}\right]$ and $y_D = \left[\begin{array}{c} y_S \\ y_S\end{array}\right]$. This may occur at hidden linear layers which may include attention and/or feed-forward layers.
\end{itemize}

The expanded weight parameter is formed by stacking the original pretrained matrix in both rows and columns and normalizing the values by $\frac{1}{n}$, where $n$ is the expansion factor in the input dimension. The expanded bias vector is created by repeating the original bias values $n$ times. This formulation ensures that the outputs of the expanded linear layer are cloned versions of the original linear layer's outputs. More specifically:

\begin{itemize}
    \item \textbf{Case 1:} We initialize $W_D = \transposeStackedVector{\frac{W_S}{2}+\eta_1}{\frac{W_S}{2}-\eta_1}$ and $b_D = b_S$, where $\eta_1$ is a random tensor with reasonable magnitude. We then have: $$y_D = W_D x_D + b_D = \transposeStackedVector{\frac{W_S}{2}+\eta_1}{\frac{W_S}{2}-\eta_1} \stackedVector{x_S}{x_S} + b_S = y_S$$
    \item \textbf{Case 2:} We initialize $W_D = \stackedVector{W_S}{W_S}$ and $b_D = \stackedVector{b_S}{b_S}$. We then have: $$y_D = W_D x_D + b_D = \stackedVector{W_S}{W_S} x_S + \stackedVector{b_S}{b_S} = \stackedVector{W_Sx_S+b_S}{W_Sx_S+b_S} = \stackedVector{y_S}{y_S}$$
    \item \textbf{Case 3:} We initialize $W_D = \stackedMatrix{\frac{W_S}{2}+\eta_1}{\frac{W_S}{2}-\eta_1}{\frac{W_S}{2}+\eta_2}{\frac{W_S}{2}-\eta_2}$ and $b_D = \stackedVector{b_S}{b_S}$, where $\eta_1$ and $\eta_2$ are a random tensors with reasonable magnitudes. We then have: $$y_D = W_D x_D + b_D = \stackedMatrix{\frac{W_S}{2}+\eta_1}{\frac{W_S}{2}-\eta_1}{\frac{W_S}{2}+\eta_2}{\frac{W_S}{2}-\eta_2} \stackedVector{x_S}{x_S} + \stackedVector{b_S}{b_S} = \stackedVector{y_S}{y_S}$$
\end{itemize}

% This technique allows us to transfer the parameters from a small linear layer to a large linear layer while preserving the layer's functionality.

\noindent{\bf Cloning Attention Layers.} 
When cloning attention layers, there are two possibilities to expand a multi-head attention:

\begin{itemize}
    \item \textbf{Expanding the dimension of each attention head:} When increasing the head dimension, each of the query/key/value matrices can be treated as individual linear layers and expanded as explained in Figure~\ref{fig:linear_cloning}. Let $q_S$ and $k_S$ represent the query and key values in the small network. Then the corresponding query and key values in the expanded network would be:
\[
q_D = \stackedVector{q_S}{q_S} \quad \text{and} \quad k_D = \stackedVector{k_S}{k_S}
\]

The attention computed in the small network is:
\[
a_S = \frac{q_S k_S^T}{\sqrt{d}}
\]

In the expanded layer, the attention is computed as:
\[
a_D = \frac{q_D k_D^T}{\sqrt{2d}} = \frac{q_S k_S^T + q_S k_S^T}{\sqrt{2d}} = \sqrt{2} a_S
\]

To make $a_D$ equal to $a_S$, we should scale the query value by $\frac{1}{\sqrt{2}}$. More generally, the expanded query weights should be scaled by $\sqrt{\frac{d_{\text{S}}}{d_{\text{D}}}}$,
where \( d_{\text{S}} \) and \( d_{\text{D}} \) are the head dimensions in the original and extended layers, respectively.

    \item \textbf{Expanding the number of attention heads:} This case is straightforward. We can simply duplicate the attention heads.
\end{itemize}

In both cases, the fully connected layer that follows the attention layer will also be expanded to increase the hidden representation's dimensionality.

\noindent{\bf Cloning Layer Norm}. 
let $x_S$ be a hidden representation vector in the small network. Applying Layer Norm over this vector computes the following:
$$l(x_S) = \frac{x_S-\mathbb{E}(x_S)}{\sqrt{var(x_S)+\epsilon}}\cdot \gamma_S + \beta_S$$
When cloning layer norm, we expand the affine parameters (if any) as $\beta_D = \stackedVector{\beta_S}{\beta_S}$ and $\gamma_D = \stackedVector{\gamma_S}{\gamma_S}$. We then have:
$$l(x_D) = \frac{\stackedVector{x_S}{x_S}-\mathbb{E}(\stackedVector{x_S}{x_S})}{\sqrt{var(\stackedVector{x_S}{x_S})+\epsilon}}\cdot \stackedVector{\gamma_S }{\gamma_S }+ \stackedVector{\beta_S }{\beta_S} = \stackedVector{l(x_S)}{l(x_S)} $$
In the above derivation, we used the fact that $\mathbb{E}(\stackedVector{x_S}{x_S})=\mathbb{E}(x_S)$ and $var(\stackedVector{x_S}{x_S}) = var(x_S)$. In general, repeating the weights and biases in the layer norm $n$-times will ensure that the output of the expanded layer norm is a cloned version of the output from the original layer norm. Similar argument is true for batch normalization, RMS normalization, and group normalization.

% \TODO
\noindent{\bf Cloning Positional Embedding Layers}. 
For positional embedding, we need to define the $n$-times cloned equivalents. Let $P_S(x_S, i) \in \mathbb{R}^d$ denote the positional embedding of a pretrained small network. The $n$-times cloned positional embedding is defined as follows:
\[ P_D(X_D, i) =
    \left[
    \begin{array}{c}
        P_S(x_S, i) \\
        \vdots \\
        P_S(x_S, i)
    \end{array}
    \right]
\]
In essence, the positional embedding of the expanded network is created by repeating the positional embedding of the small network $n$ times. In our codebase, we define Pytorch equivalents of the expanded positional embedding layers when necessary.

\section{Architectures and Training Details}~\label{app:training}

The architectures of our etudied networks are summarized in Table~\ref{tab:arcs}. Among the target models, OPT-1.3B and Pythia-1.4B are already available through HuggingFace, providing us with a good baseline for comparison. OLMO-2.9B was not trained by the authors of~\citep{groeneveld267365485olmo}, and we are the first to train and evaluate it. We obtain the weight checkpoints of the base models from the HuggingFace repositories, except for OPT-350M, for which we train our own base model with 30B tokens. This is because the HuggingFace OPT-350M model has extra linear layers after the embedding layer and before the unembedding layer, which the target OPT-1.3B model does not have. With these benchmarks, we emulate three different scenarios:

\begin{itemize}
    \item \textbf{OPT}: The training dataset is the same for both the base and target model. The base model is trained with a relatively small number of tokens (30B).
    \item \textbf{Pythia}: The dataset used for training the base model (Pile) is not available to train the target model. We use a different dataset (DOLMA) for training the target model. The base model was trained with a moderate number of tokens (\~250B).  
    \item \textbf{OLMO}: The training dataset is the same for both the base and target model. The base model is trained with a large number of tokens (2.4T).
\end{itemize}

\begin{table}[ht]
\centering
\caption{Summary of base and target model architectures.}\label{tab:arcs}
\begin{tabular}{cccccc}
\multicolumn{2}{c}{\textbf{Model}} & \textbf{\#L} & \textbf{\#H} & \textbf{d\textsubscript{model}} & \textbf{d\textsubscript{FFN}} \\ \hline
\textbf{base}      & OPT 350M      & 24           & 16           & 1024                  & 4098                \\
\textbf{target}    & OPT 1.3B      & 24           & 32           & 2048                  & 8192                \\ \hline
\textbf{base}      & Pythia 410M   & 24           & 16           & 1024                  & 4098                \\
\textbf{target}    & Pythia 1.4B   & 24           & 32           & 2048                  & 8192                \\ \hline
\textbf{base}      & OLMO 1B       & 16           & 16           & 2048                  & 16384              \\
\textbf{target}    & OLMO 2.9B     & 16           & 32           & 4096                  & 16384              
\end{tabular}
\end{table}

\noindent{\bf Dataset.} For all experiments, we use the DOLMA dataset provided by the authors of~\citep{groeneveld267365485olmo}. This dataset includes several open-source datasets and totals up to 2.4 trillion tokens. However, our training jobs do not process this many tokens due to the extensive cost. To ensure fair representation of all sub-datasets within DOLMA, we shuffled the data shards. The seed for random shuffling is kept the same across all our experiments to eliminate the impact of data ordering on our conclusions.

\noindent{\bf Training Parameters.} For all of our experiments, we use the AdamW optimizer with a weight decay of 0.05, $\beta_1=0.9$ and $\beta_2=0.999$. We use gradient accumulation with 16 steps to increase our effective batch size and the zero\_2 gradient update algorithm~\citep{rajbhandari2020zero} to reduce memory footpring. We apply a learning rate warm-up over 25,000 iterations to reach the maximum learning rate. Afterward, we decay the learning rate to 1/10th of its value until 2,500,000 iterations, after which the learning rate is kept constant. Our models are trained on 64 GPUs with varying batch sizes, context sizes, and learning rates summarized in Table~\ref{tab:training}.

\begin{table}[ht]
\centering
\caption{Training hyperparameters.}\label{tab:training}
\begin{tabular}{ccccc}
\textbf{Model}       & \textbf{Batch Size} & \textbf{Context Size} & \textbf{Max LR} & \textbf{\begin{tabular}[c]{@{}c@{}}Average Tokens\\ Per Iteration\end{tabular}} \\ \hline
\textbf{OPT 1.3B}    & 2                   & 1024                  & 1.5E-4          & 65K                                                                             \\ \hline
\textbf{Pythia 1.4B} & 2                   & 1024                  & 1.5E-4          & 65K                                                                             \\ \hline
\textbf{OLMO 2.9B}   & 2                   & 2048                  & 3E-4            & 82K                                                                            
\end{tabular}
\end{table}

% Optionally include supplemental material (complete proofs, additional experiments and plots) in appendix.
% All such materials \textbf{SHOULD be included in the main submission.}

%%%%%%%%%%%%%%%%%%%%%%%%%%%%%%%%%%%%%%%%%%%%%%%%%%%%%%%%%%%%

% \newpage

\end{document}